\documentclass{article}
\usepackage{arabtex}
\newcommand{\persianChehelSalegi}[1]{{\setfarsi\novocalize\<^chl sAlgy>}}
\usepackage{arxiv}
\usepackage{amsmath}

\usepackage{pgfplots}
\pgfplotsset{width=10cm, compat=newest}

\title{An Unsupervised Language-Independent Entity Disambiguation Method and its Evaluation on the English and Persian Languages}

\author{
    Majid Asgari-Bidhendi\\
    School of Computer Engineering\\
    Iran University of Science and Technology\\
    Tehran, Iran\\
    \texttt{majid.asgari@gmail.com}\\
    \And
    Behrooz Janfada\\
    School of Computer Engineering\\
    Iran University of Science and Technology\\
    Tehran, Iran\\
    \texttt{behrooz.janfada@gmail.com}\\
    \And
    Amir Havangi\\
    School of Computer Engineering\\
    Iran University of Science and Technology\\
    Tehran, Iran\\
    \texttt{havangi@yahoo.com}\\
    \And
    Sayyed Ali Hossayni\\
    School of Computer Engineering\\
    Iran University of Science and Technology\\
    Tehran, Iran\\
    \texttt{hossayni@iran.ir}\\
    \And
    Behrouz Minaei-Bidgoli\\
    School of Computer Engineering\\
    Iran University of Science and Technology\\
    Tehran, Iran\\
    \texttt{b\_minaei@iust.ac.ir}\\
}
\renewcommand{\shorttitle}{ULIED}
\begin{document}
\maketitle

\begin{abstract}
Entity Linking is one of the essential tasks of information extraction and natural language understanding. Entity linking mainly consists of two tasks: recognition and disambiguation of named entities. Most studies address these two tasks separately or focus only on one of them. Moreover, most of the state-of-the -art entity linking algorithms are either supervised, which have poor performance in the absence of annotated corpora or language-dependent, which are not appropriate for multi-lingual applications. In this paper, we introduce an Unsupervised Language-Independent Entity Disambiguation (ULIED), which utilizes a novel approach to disambiguate and link named entities. Evaluation of ULIED on different English entity linking datasets as well as the only available Persian dataset illustrates that ULIED in most of the cases outperforms the state-of-the-art unsupervised multi-lingual approaches.
\end{abstract}


\section{Introduction}
\label{sec:intro}
In this section, we first present a general introduction to entity linking (EL). We then introduce the knowledge bases as an essential requirement for entity linking task and then discuss the general steps of it. At the end of this section, we outline the general structure of this paper.
\subsection{Entity Linking}
Entity Linking (EL) is the task of linking a set of entities mentioned in a text to an external dataset. Entity linking plays an essential role in text analysis, information extraction, question answering, text understanding, and recommender systems~\cite{yan2017entity}. It also allows users to know about the background knowledge of entities in the text~\cite{han2011collective}. However, there are two types of ambiguities which make this task challenging. Firstly, entities may have different names, even in a single document. For example, the name of a person can appear in the text as the first name, last name, or nickname.
EL should link all of these names to a single entity in the knowledge base. Secondly, different entities may have the same name, but the entity linking system should be able to refer them to various entities from the knowledge base. Therefore, information about entities is crucial in choosing the correct entities~\cite{ganea2016probabilistic,han2011collective,shen2014entity}.

With few exceptions, most of entity linking methods separately address the mention detection (also known as entity detection or named entity detection) and entity disambiguation stages~\cite{kolitsas2018end}. Our approach also focuses on entity disambiguation. The proposed approach is an Unsupervised Language-Independent Entity Disambiguation method, dubbed as ULIED.

\subsection{Knowledge Base}
A knowledge base is one of the fundamental components in entity linking systems. Generally, the KB consists of a set of entities, information, semantic categories, and the relationship between entities. Knowledge bases used in EL systems should have features such as public availability, machine readability, persistent identifiers, and credibility~\cite{taufer2017named}. There are currently several knowledge bases for EL systems such as DBpedia~\cite{auer2007dbpedia}, YAGO\footnote{Yet Another Great Ontology }~\cite{suchanek2007yago}, Freebase~\cite{bollacker2008freebase}, and Probase~\cite{wu2012probase}. In the case of low-resource languages, cross-lingual methods are used if there is no knowledge base with the above characteristics~\cite{zhou2020improving}.
Fu et al.~\cite{fu2020design} showed that cross-lingual methods are heavily dependent on Wikipedia and only work well on Wikipedia texts. These methods perform poorly in non-Wikipedia texts and require outside-Wikipedia cross-lingual resources to improve their performance. This study employs FarsBase~\cite{asgari2019farsbase}, which is the first multi-source KB especially designed for the Persian language and includes more than 500,000 entities with 25 million relations between them. FarsBase can provide various information such as locations, persons, and organizations.

\subsection{Entity Linking Steps}
Generally, the EL process includes four subtasks, which are consistent with most of both supervised and unsupervised EL systems. The first step, Mention Detection (MD), is the operation of specifying named entities in the input natural language raw text. The last three steps can be grouped as Entity Disambiguation (ED) subtask. ED is the operation of the disambiguation of a named entity using a set of candidate entities and then linking it to a knowledge base.

\paragraph{Mention Detection} An MD algorithm captures the raw text and specifies the place of occurrence of named entities at the output. Most studies~\cite{pershina2015personalized,ran2018attention,hoffart2011robust} in EL employed existing algorithms, provided by the other researches, for MD, and address the other three modules.

\paragraph{Candidate Entity Generation} In this step, the system proposes a set of candidate entities for every entity mentions in the text provided by the previous step~\cite{shen2014entity,wu2018entity}. In this regard, most studies~\cite{shen2014entity,han2009nlpr_kbp,chen2010cuny,xu2018unsupervised} use features such as redirect pages, disambiguation pages, and hyperlinks in Wikipedia (or other knowledge bases and resources), to make a name dictionary for each entity mention with the aim of mapping the entity mention to a set of candidate entities. 

\paragraph{Candidate Entity Ranking} In most cases, candidate entity generation modules, generates multiple candidate entities for a single mention. Therefore, candidate entities should be ranked by the EL system to find the most likely entity from the knowledge base\cite{taufer2017named}. The EL system can use two types of features for ranking candidates entities: Context-Independent Features and Context-Dependent Features\cite{shen2015ranking}. In the literature, the term \textit{``phase entity disambiguation''}~\cite{cucerzan2007large,dredze2010entity,yamada2016joint} has the same meaning as candidate entity ranking. Additionally, both supervised and unsupervised methods can be used to achieve the results. Supervised ranking methods depend on the annotated training dataset, where its data annotation should be done manually. In the case of low-source languages, such a resource is not available, and alternative solutions are needed. For example, Kile et al.~\cite{klie2020zero} have provided a novel domain-agnostic Human-In-The-Loop (HITL) annotation approach which uses recommenders that suggest potential concepts and adaptive candidate ranking to speed up the overall annotation process and make it less tedious for users. NERank+ proposed by Wang et al.~\cite{wang2018nerank} is another example of entity ranking approaches, which utilizes Topical Tripartite Graph, consisting of document, topic and entity nodes as well as a random walk algorithm to propagate prior entity and topic ranks based on the graph model.

\paragraph{Unlinkable Mention Prediction} In cases where entity mentions do not have any relevant entities in the knowledge base, unlinkable entity mentions are separated from other entities and tagged as NIL. Different approaches are suggested by researchers to separate unlinkable entity mentions: (1) ignoring unlinkable entity mentions~\cite{han2011collective,han2012entity,pershina2015personalized}, (2) ignoring the low-probability candidates(NIL threshold)~\cite{shen2012linden,yamada2016joint} and (3) supervised machine learning techniques~\cite{shen2014entity,taufer2017named,zhang2010nusi2r,zhang2011entity}.

The rest of this paper is organized as follows. Section~\ref{sec:related-works-sec} presents an overview study of EL, especially with a view to entity disambiguation and language independent and unsupervised approaches. In this section, we will also introduce the only available entity disambiguation dataset for the Persian language, and we will review multilingual entity linking datasets as well as entity disambiguation datasets for the English language. Section~\ref{sec:proposed-approach} describes the proposed approach for unsupervised, language-independent entity disambiguation. Experimental results and the comparison of obtained results with the baseline methods on English and Persian datasets are discussed in section~\ref{sec:results-sec}. The last section concludes this research and expresses our future work.

\section{Related Work}
\label{sec:related-works-sec}
Supervised approaches need adequate resources and are not suitable for low-resource languages. There are a limited number of EL system which are focused on multilingual strategies. This manuscript targets multilingual unsupervised entity disambiguation as the best approach for entity linking in low-resources languages. In this section, we first describe the related literature in unsupervised EL and multilingual EL and then introduce some popular existing datasets for entity disambiguation.

\subsection{Unsupervised Entity Disambiguation}
There are various algorithms proposed to perform unsupervised EL. Here we review notable studies in this field.

Some researchers~\cite{chen2010cuny,cucerzan2007large,han2009nlpr_kbp,xu2018unsupervised} used Vector Space Model (VSM)~\cite{salton1975vector} based methods for unsupervised candidate ranking. In this method, the first step is calculating the similarity between the vector representations of the entity mention and the candidate entity. The system links the candidate entity with the highest similarity to the entity mention. These methods are different in the calculation of vector similarity and vector representation~\cite{shen2014entity}. Methods have also been proposed that use a combination of different SVM methods as an ensemble. For example, Alokaili and Menai proposed an ensemble learning using SVM~\cite{alokaili2020svm}, which produces competitive performance levels compared to well-known entity annotation systems and ensemble models on different benchmark corpora.

Cucerzan~\cite{cucerzan2007large} used entity mentions and Wikipedia articles of the candidate entities to build vectors. To this end, the system will choose a candidate that maximizes vector similarity and have the same category as an entity mention. This system got 91.4\%  accuracy on a news dataset.

Chen et al.~\cite{chen2010cuny} built the entity mention and candidate entities vectors based on the Bag of Words model by using the context of their article to capture word co-occurrence information and computed the similarity between them by TF-IDF similarity. They reported 71.2\%  accuracy on the TAC-KBP2010 dataset.

Han and Zhao~\cite{han2009nlpr_kbp} used two types of similarity measures: the Wikipedia Semantic Knowledge-Based Similarity alongside Bag of Words based similarity. For generating vectors in the first similarity, the method detects Wikipedia concepts in candidate entities and context of the mentioned entity. It then computes the vector similarity of the entity mention and candidate entities using a weighted average of semantic relations between articles of Wikipedia concepts and the context of the mentioned entity. After that, these two types of similarity are merged, and the final similarity vector of the candidate entities is reported, and finally, the entity that maximizes this merged similarity is chosen. Their system achieves 76.7\%  accuracy on the TAC-KBP2009 dataset.

Xu et al.~\cite{xu2018unsupervised} applied a linking approach for medical texts and exploited name similarity, entity popularity, category consistency, context similarity, and the semantic correlation between the entity mention and candidate entities, and ranked candidate entities by combining these features. They called their ranking measure, Confidence Score. On average, their Confidence Score got about 82\%  precision on their medical dataset.

Zhang et al.~\cite{zhang2017xlink} proposed an unsupervised bilingual entity linker inspired by Han et al.~\cite{han2011collective} and Yamada et al.~\cite{yamada2016joint} researches. As we discussed before, they utilized a pre-built dictionary for the candidate generation, and after that, they used probabilistic generative methods to disambiguate the entities. Their system achieves 91.2\%  precision on the CoNLL dataset.

Pan et al.~\cite{pan2015unsupervised} used Abstract Meaning Representation (AMR)~\cite{banarescu2013abstract} to select high-quality sets of entities for their similarity measure. They claimed that their representation using AMR could capture some contextual properties which are very critical and helpful for disambiguating entities without using training data. Next, for comparing the context of the entities, they used an unsupervised graph to get final results and reported 92.12\%  precision on a dataset annotated from news and discussion forum posts.

Xie et al.~\cite{xie2020graph} proposed the graph-ranking collective Chinese entity linking (GRCCEL) algorithm, which utilizes both the structured relationship between entities in the local knowledge base and the additional background information offered by external knowledge sources. To measure similarity, they used improved weighted word2vec and improved PageRank methods. They reported the effectiveness of GRCCEL in Chinese entity linking task and demonstrated the superiority of their method over state-of-the-art methods in Chinese.

\subsection{Multi-Lingual Entity Linking}
\label{sec:multilingual-entity-linking}
Some studies target multilingual entity linking. Babelfy~\cite{moro2014entity} is one of the most distinguished studies on unsupervised multilingual EL and Word Sense Disambiguation (WSD). Moro et al. used a unified graph-based approach to EL and WSD based on a loose identification of candidate meanings coupled with the densest subgraph heuristic, which selects high-coherence semantic interpretations.

Hoffart et al. proposed the AIDA~\cite{hoffart2011robust} system which provides an integrated NED method using popularity, similarity, and graph-based coherence, and includes robustness tests for self-adaptive behavior. Later, they extended their approach~\cite{hoffart2012kore} presenting a novel notion of semantic relatedness between two entities represented as sets of weighted (multi-word) keyphrases, with consideration of partially overlapping phrases. This measure improves the quality of prior link-based models and also eliminates the need for (usually Wikipedia-centric) explicit interlinkage between entities.

Usbeck et al.~\cite{usbeck2014agdistis} present AGDISTIS, a knowledge-base-agnostic approach for named entity disambiguation. Their approach combines the Hypertext-Induced Topic Search (HITS) algorithm with label expansion strategies and string similarity measures. They extended their AGDISTIS to a multilingual approach named MAG\cite{moussallem2017mag}.

Rosales et al.~\cite{rosales2018voxel} introduce VoxEL as a benchmark dataset for multilingual Entity Linking
including German, English, Spanish, French and Italian languages based on 15 news articles
from VoxEurop, a multilingual newsletter, totaling 94 sentences.
The study compares 15 multilingual entity linkers with General Entity Annotation Benchmark Framework (GERBIL)~\cite{usbeck2015gerbil}:
KIM~\cite{popov2004kim},
TagME~\cite{ferragina2010tagme},
SDA~\cite{charton2011automatic},
ualberta~\cite{guo2012ualberta},
HITS~\cite{fahrni2012hits},
THD~\cite{dojchinovski2012recognizing},
DBpedia Spotlight~\cite{mendes2011dbpedia,daiber2013improving},
Wang-Tang~\cite{wang2013boosting},
AGDISTIS~\cite{usbeck2014agdistis},
Babelfy~\cite{moro2014entity},
FREME~\cite{sasaki2016chainable},
WikiME~\cite{tsai2016cross},
FEL~\cite{pappu2017lightweight},
FOX~\cite{speck2017ensemble} and
MAG~\cite{moussallem2017mag}
and checks the availability of entity recognition, having a demo, having an API and availability of source codes for each system.

DBpedia Spotlight proposed by Mendes et al.~\cite{mendes2011dbpedia} is a system for automatically annotating text documents with DBpedia URIs. Their algorithm uses the same four-step approach and VSM and TF-IDF similarity measure, which is described earlier.

\section{Datasets}
\label{subsec:dataset-sec}
In this section, we present a review of some popular entity disambiguation datasets, for the English language, which is used in the evaluation of this research and the only published dataset for the Persian language.

\subsection{ACE2004}
In the 2004 Automatically Content Extracting (ACE) technology evaluation, the ACE 2004 Multilingual Training Corpus contains all English, Arabic, and Chinese education data. This collection includes various types of data annotated for organizations and connections, and a Linguistic Information Consortium has been formed with the help of the ACE Program with the additional support of the DARPA TIDES program.

\subsection{AIDA/CoNLL}
This dataset contains assignments of entities to the mentions of named entities annotated for the original CoNLL 2003 entity recognition task~\cite{hoffart2011robust}. This dataset consists of proper noun annotations for 1393 Reuters newswire articles. All these proper nouns are hand-annotated with corresponding entities in YAGO2. Two experts disambiguated each mention, and in case of conflict, another expert resolved the conflict.

\subsection{AQUAINT}
AQUAINT Corpus~\cite{crouch2005aquaint}, Linguistic Data Consortium (LDC) catalog number LDC- 2002T31 and ISBN 1-58563-240-6 consists of English-language newswire text data from three sources: the Xinhua News Service (People's Republic of China), the New York Times News Service, and the Associated Press Worldstream News Service. It was prepared for the AQUAINT Project by the LDC and will be used by the National Institute of Standards and Technology (NIST) in official benchmark evaluations.

\subsection{DBpediaSpotlight}
DBpedia Spotlight~\cite{mendes2011dbpedia} is a system where text documents with DBpedia URIs are automatically annotated. DBpedia Spotlight enables users to configure annotations to their specific needs utilizing DBpedia Ontology and quality measures such as prominence, topical relevance, contextual ambiguity, and confidence in disambiguation.

\subsection{KORE50}
The goal of KORE50~\cite{hoffart2012kore} is to stress testing EL systems using highly ambiguous mentions using hand-crafted sentences via difficult disambiguation tasks.
KORE is a new notion of entity relatedness, based on the overlap of two sets of keyphrases, e.g., partial matches of 2 sentences.

\subsection{KORE 50-DYWC}
An excess of different evaluation data sets relies on either Wikipedia or DBpedia. Noullet et al.~\cite{noullet2020kore} have recently extended KORE50, to not only accommodate EL tasks for DBpedia, but also for YAGO, Wikidata, and Crunchbase.

\subsection{IITB}
IITB~\cite{kulkarni2009collective} was established in 2009 and had the highest corporate entity/document density.
It is a list of ground truths (called "IITB") using an annotation system based on a browser. Manual annotation documents were gathered from the links to popular websites belonging to a handful of domains that included sports, culture, science and technology, and education. The annotations can be found in the public domain.

\subsection{N3-Reuters-128}
N3 Reuters-128~\cite{roder2014n3} includes 128 news articles sampled randomly from the Reuters-21578 news articles and manually annotated by domain experts.

\subsection{N3-RSS-500}
N3 RSS-500~\cite{roder2014n3} consists of 1,457 RSS feeds scraped from a list containing all major newspapers around the world and a wide range of topics. Domain experts manually annotated the corpus.
The RSS list was compiled using a 76-hour crawl, leading to a corpus of approximately 11.7 million sentences. By render selecting 1\% of the contained words, a subset of this corpus was generated.

\subsection{ERD2014}
The ERD2014~\cite{carmel2014erd} dataset is constructed for the 2014 Entity Recognition and Disambiguation Challenge (ERD'14), which took place from March to June 2014 and was summarized in a dedicated workshop at SIGIR 2014. The ERD challenge's main goal was to promote research in recognition and disambiguation of entities in unstructured text.
For the short-text track, the dataset was built by sampling 500 queries from a commercial search engine's query log to form a development set and 500 queries for the test set. The average query length was four words per query. For the long-text track, the dataset was built by sampling 100 web pages for the development and 100 web pages for the test set. Also, all HTML tags from the web pages were stripped, and various heuristics were applied to extract the main content from each document. Particularly,
boilerplate content from header and side panes were removed. Among all documents, 50\%
were sampled from general web pages; the remaining 50\% were news articles from msn.com.

\subsection{MSNBC}
Silviu Cucerzan launched MSNBC dataset~\cite{cucerzan2007large} in 2007. The data set contains unique SF media reports and a distinctive lexicalization.

\subsection{ParsEL-Social}
To evaluate ULIED and competing methods in Persian, we used ParsEL-Social Dataset~\cite{fakhrian2019parsel}, which is introduced in the conference version of this paper. ParsEL-Social is constructed from social media contents derived from 10 Telegram channels in 10 different categories: sport, economics, gaming, general news, IT news, travel, art, academic, entertainment, and health.
To create this dataset, firstly, entity mentions are automatically identified in raw text, and a list of candidates is created based on redirect and disambiguation links of that entity mention in Wikipedia. Then the text with identified entity mentions and candidate lists is given to a Persian linguistics expert. The expert either selects one of the candidates for each entity. In some cases, a mention must be linked to an entity. However, the right entity is not found in the candidates due to an error in the automatic candidate generation algorithm. The expert may add it to the candidate list and selects it as the right link. It should be noted that the automatic candidate generator only appears as an expert helper.

\section{Proposed Approach}
\label{sec:proposed-approach}
In this section, we describe the architecture of our unsupervised language independent entity disambiguation system (ULIED) and propose our new approach in which entity mentions of input text are disambiguated and linked to Wikipedia. We first look at the architecture as a general system and then describe each of its modules separately. 

\subsection{Unsupervised Language-Independent Entity Disambiguation System (ULIED) Architecture}
\begin{figure}[ht]
    \centering
	\includegraphics[width=\linewidth]{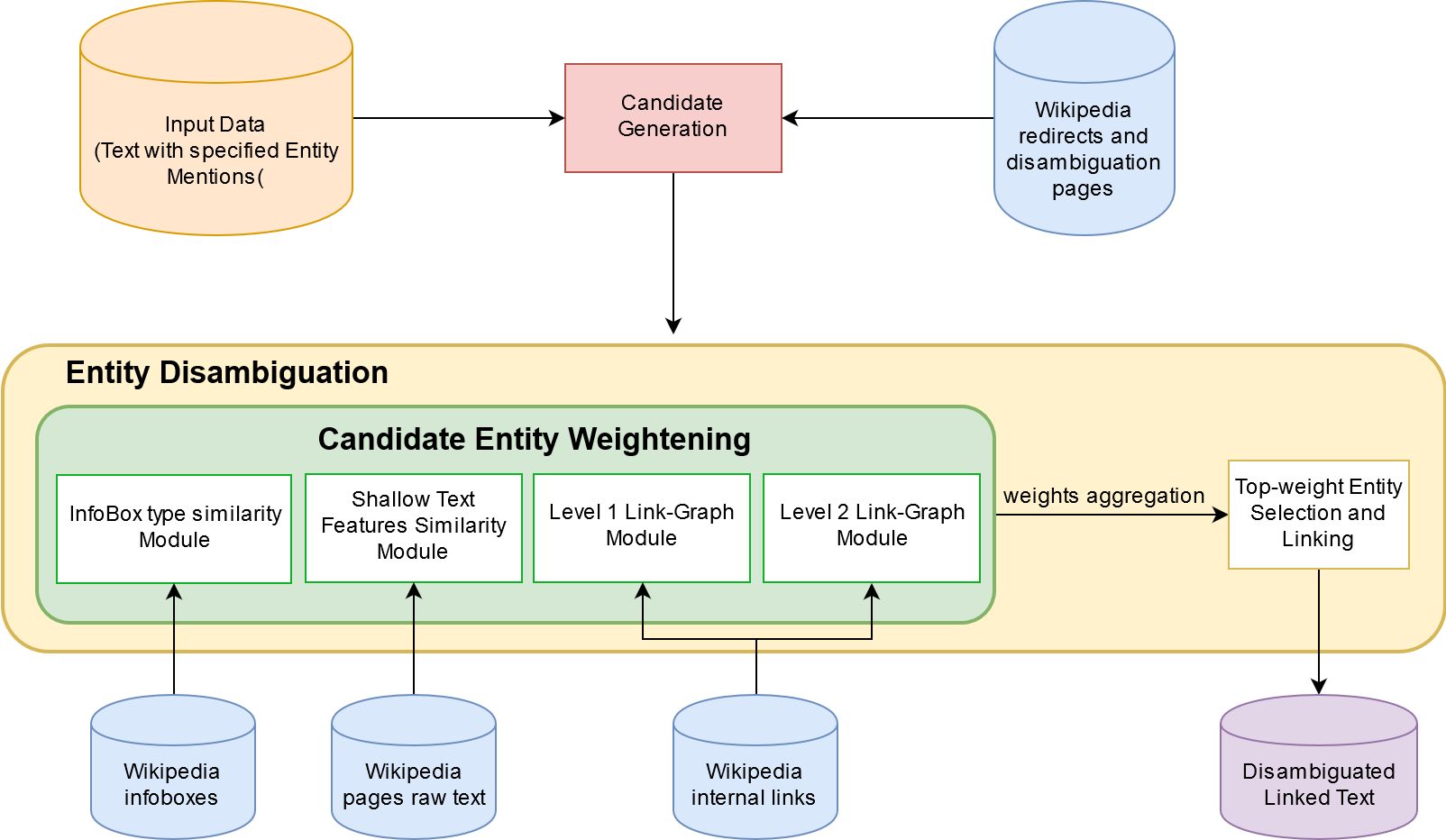}
	\caption{The architecture of unsupervised language-independent entity disambiguation system (ULIED)}
	\label{fig:Arch}
\end{figure}

The ULIED system implemented in a pipeline architecture. It is assumed that the input text only specifies the entity mentions, and the system at the output should disambiguate these entity mentions and link them to Wikipedia. To this end, a list of candidates is first generated in the Candidate Generation module for each entity mention. Four candidate entity weighting modules are used in the ED component; two context-dependent modules and two context-independent modules. Figure \ref{fig:Arch} shows a block diagram of the architecture and data flow of ULIED. 

In this architecture, in the Candidate Generation module, candidates are generated for all entity mentions of the text using Wikipedia redirects and disambiguation pages. The output of this module is a set of entity mentions and several candidates for each entity mention.

Entity Disambiguation Component consists of five modules; four are grouped as Candidate Entity Weightening sub-system and one named Top-weight Entity Selection and Linking Module. We utilize both of the context-dependent and context-independent features in the ranking step. Context-dependent features rely on the context where entity mention appears, but context-independent features are independent of context and rely on entity mention and candidate entities\cite{shen2014entity}. These modules are described below.

\subsubsection{Infobox type similarity Module}
In this module, we examine whether the type of candidate can match the entity's context by the type of candidate entity derived from Wikipedia's Infobox and the context surrounding the original entity.

Some entities have a very generic name that may cause a high level of ambiguity. For instance, \persianChehelSalegi~ (``At the age of 40'') is an Iranian movie while it can be part of a general sentence, e.g., \  ``Vahid died at the age of 40''. Such names are widespread in artworks (e.g., \ movies or books) and a limited number of the other specialized classes. To improve the disambiguation process, we will look for more evidence in the context using a hand-made reference list if the candidate entity belongs to individual classes. Considering the above example, ``At the age of 40'', the surrounding context containing phrases such as a director, artist, channel, cinema, ticket, and movie is required. Otherwise, the algorithm multiplies the candidate's real rate by a predefined constant number between 0 and 1 based on each case.

\subsubsection{Shallow Textual Similarity Module}
In this module, using the surface properties of the text of the document that contains a given entity mention and corresponding Wikipedia page of its candidates, we perform a vector similarity to determine the degree of similarity between each candidate and its corresponding entity mention.
\subsubsection{Level 1 and Level 2 Link-Graph Module}

Previous research has shown that internal links between Wikipedia pages can be a useful feature for examining and measuring the semantic relevance of concepts~\cite{zhu2019efficient}. The main idea behind the Link-Graph modules is that candidates with more Wikipedia internal links to other candidates are more likely to be suitable candidates for disambiguation. 

In the following, we will explain this method in more detail. Given the candidate list $CL$ including all candidates of all mentions in the context, for each candidate $c_i \in CL$, we create a list $LLC^1_i$ including all candidates which are linked in the corresponding Wikipedia article for $c_i$ and their frequencies in the article, $(link_{ij},count_{ij})$. For example, the Wikipedia article of ``Saadi'' contains ten links to ``Shiraz'' article, four links to ``Persian'', 12 links to ``Poet'', and so on. The list $LLC^1_{Saadi}$ will be:

\begin{equation}
[(Shiraz,10), (Persian,4), (Poet,12), ...]. 
\end{equation}

In the next step, we assign a weight to each $c_i$ of $CL$:
\begin{equation}
w_{c_i} = \sum_{(link_{ij},count_{ij}) \in LLC^1_i} count_{ij} \times e_{ij}
\end{equation}
and $e_{ij}$ is
\begin{equation}
\ e_{ij} =\begin{cases}
1 &  \mathrm{if} ~ link_{ij} \in CL \\
0 &  \mathrm{if} ~ otherwise
\end{cases}
\end{equation}

In the last step, for each mention, we choose the entity with the highest weight for the mention.

The Level 1 Link-Graph Module only counts the number of links per candidate and accordingly assigns a weight to each candidate. The Level 2 Link-Graph Module is not limited to first-level links. Instead, we create a $LLC^2$ list which equals to $LLC^1$ of the entity + $LLC^1$ of entities which are linked to each of them. In the previous example about ``Saadi'', we add all of the links of ``Shiraz'', ``Persian'' and ``Poet'' article to $LLC^2$ of Saadi.

To accelerate this phase, we have created a cache $LLC$ for all of the articles in Wikipedia dump, and the system can fetch the list for each member of list $CL$ instantly from the cache.

Suppose the following text as the input:\
``Saadi was born in the city of Shiraz.''

\paragraph{Level-1 graph formation:} Suppose ``Saadi'' has three candidate entities: $A1$, $A2$, and $A3$, and the ``city'' can also refer to two entities: $B1$ and $B2$. ``Shiraz'' also has four ambiguities: $C1$, $C2$, $C3$, and $C4$. The rest of the words have no candidate entity. In this example, a graph is formed that has the following nodes:

$A1, A2, A3, B1, B2, C1, C2, C3, C4$

If the Wikipedia page of a node has $n$ internal links to another node's Wikipedia page, an edge of value $n$ is created between the two nodes.

In the above example, ``Saadi'' has two candidates. Suppose the Wikipedia page of candidate $A1$ is linked to the ``Shiraz'' Wikipedia page ($C2$), and candidate $A2$ is not linked to this page. Considering that the city of Shiraz ($C2$) is one of the candidates for the word ``Shiraz'' in the text and has appeared in the graph, the score of the first candidate, ``Saadi'' ($A1$), will be higher than the second candidate ($A2$).

\paragraph{Level-2 graph formation:} In the previous example, suppose that Saadi's ($A1$) candidate is linked to the other four entities $D1$, $D2$, $D3$, and $D4$. In the second level graph, these entities are also added to the graph. The adding operation will be done for all other level-1 nodes ($A2$, $A3$, $B1$, $B2$, $C1$, $C2$, $C3$, $C4$). The level-2 graph will always be more crowded, but the criterion for forming edges is the same as the previous graph: the internal Wikipedia link between the entities' pages.

$A1, A2, A3, B1, B2, C1, C2, C3, C4, D1, D2, D3, D4, ....$ (links of all other candidates)

Suppose the previous example with some differences. ``Saadi'' has two candidates $A1$ and $A2$. Suppose that neither candidate $A1$ nor candidate $A2$ has a link to the Shiraz Wikipedia page ($C2$). However, $A1$ is linked to another Wikipedia page $D2$, and $D2$ is linked to $C2$. $A2$ has not any connection to $C2$, even with one node in between. In this example, $C2$ exists in the level-2 graph, and the score of the first candidate $A1$ will be higher than the second candidate $A2$.

\subsubsection{Top-weight Entity Selection and Linking Module}
In this module, we first multiply all the weights of the candidates obtained from the previous four modules, and then we sort all the candidates according to the final number of these results, and we consider the highest weighted candidate as the selected candidate. The link is to the Wikipedia page for this candidate. As such, the entity has been disambiguated. Finally, the system links the candidate entity with the highest score to the entity mention. Other entities will be added to the entity mention's ``ambiguity-list'' to persist the rejected candidates for possible future applications such as error checking. 
After candidate generation and ranking, the NIL threshold method is used for unlinkable mention prediction. In this method, if the score of the top-ranked candidate entity is lower than the pre-defined threshold, the entity mention will be tagged as NIL, and the system will add all of the candidate entities to the ambiguity-list.

\section{Experimental Results and Evaluation}
\label{sec:results-sec}
We implemented our proposed method, ULIED, on the Persian and English languages using existing datasets and compared the evaluation results with other state-of-the-art multilingual unsupervised methods, Babelfy, and DBpedia SpotLight. We then analyze these results and evaluations.

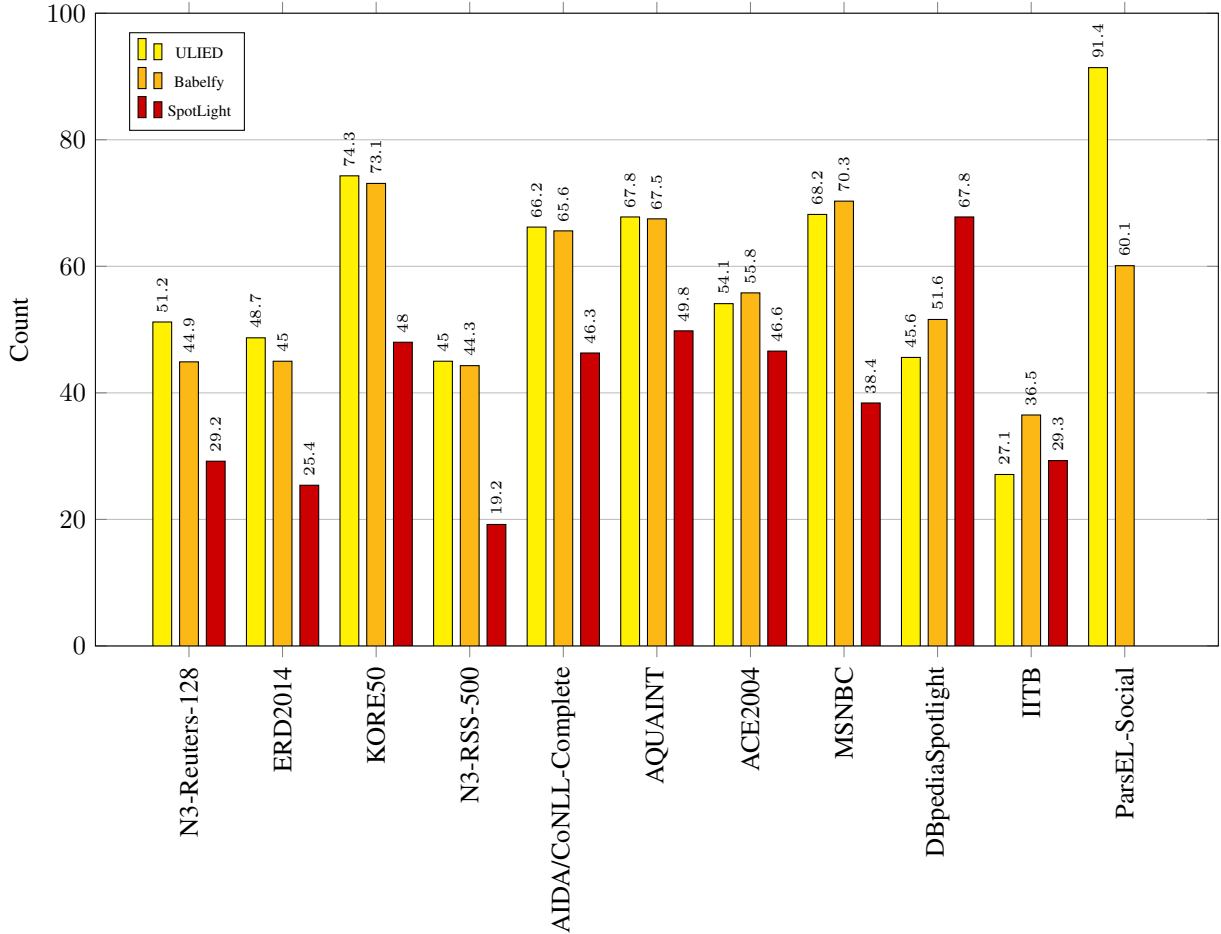
\begin{figure*}[ht]
\begin{tikzpicture}
\begin{axis}[
    ybar=3pt,
    bar width=0.25cm,
    height=10cm,
    width=1\textwidth,
    legend pos=north west,
    legend style={legend columns=1, font=\tiny},
    scaled ticks=false,
    ymin=0.0,
    ymax=100,
    ylabel={Count},
    ymajorgrids,
    symbolic x coords={N3-Reuters-128,ERD2014,KORE50,N3-RSS-500, AIDA/CoNLL-Complete,AQUAINT,ACE2004,MSNBC,DBpediaSpotlight,IITB,ParsEL-Social},
    xtick=data,
    x tick label style={rotate=90},
    nodes near coords,
    nodes near coords align={vertical},
    every node near coord/.append style={rotate=90, anchor=west, font=\tiny}
    ]

\addplot [fill=yellow] coordinates {(N3-Reuters-128,51.2)(ERD2014,48.7)(KORE50,74.3)(N3-RSS-500,45)(AIDA/CoNLL-Complete,66.2)(AQUAINT,67.8)(ACE2004,54.1)(MSNBC,68.2)(DBpediaSpotlight,45.6)(IITB,27.1)(ParsEL-Social,91.4)};
\addplot [fill=yellow!70!red] coordinates {(N3-Reuters-128,44.9)(ERD2014,45)(KORE50,73.1)(N3-RSS-500,44.3)(AIDA/CoNLL-Complete,65.6)(AQUAINT,67.5)(ACE2004,55.8)(MSNBC,70.3)(DBpediaSpotlight,51.6)(IITB,36.5)(ParsEL-Social,60.1)};
\addplot [fill=red!80!black] coordinates {(N3-Reuters-128,29.2)(ERD2014,25.4)(KORE50,48)(N3-RSS-500,19.2)(AIDA/CoNLL-Complete,46.3)(AQUAINT,49.8)(ACE2004,46.6)(MSNBC,38.4)(DBpediaSpotlight,67.8)(IITB,29.3)};
\legend{ULIED,Babelfy,SpotLight} 
\end{axis}
\end{tikzpicture}
\caption{Evaluation of entity disambiguation results using the proposed method ULIED on 10 different datasets and comparison with DBpedia-SpotLight and BabelFy as baseline algorithms. Results are reported according to the F1 measure.}
\label{result_chart}
\end{figure*}

We evaluate ULIED on ParsEL-Social dataset, and the results are reported in Figure \ref{result_chart}. 

To evaluate ULIED on the English language, we use ten different datasets that are widely used in entity linking evaluations, which are introduced in Section \ref{subsec:dataset-sec}.

To evaluate the system, we input the documents of each dataset with the spanning of each mentions within the text in the NIF standard format. Giving the position of each mentions in the text, ULIED selects some candidate Wikipedia articles for the mention and performs the ED phase and links the mention to an entity. If there is no candidate entity for the mention, or confidence value of all entities is under a threshold value, ULIED does not link the mention to any entity.

Rosales et al.~\cite{rosales2018voxel} reports the results of some state-of-the-art unsupervised language-independent EL systems on their multilingual dataset and report the superiority of Babelfy and DBpedia SpotLight on the other systems.

Figure \ref{result_chart} depicts the results of ULIED with DBpedia Spotlight and Babelfy based on micro F1-measure. The results show ULIED outperforms Babelfy in 5 datasets. Spotlight records the best performance only in the DBpediaSpotlight dataset. Therefore the performance of ULIED in the English language is comparable with other multilingual approaches and outperforms them in the Persian language.

Moreover, it is notable that the size of Wikipedia articles in the Persian language is almost 500,000 (i.e., One-twentieth of English), and consequently the number of candidates for each mention, in the Persian, is very smaller in comparison to mentions in the English language. This fact causes ULIED F-score to be meaningfully more than its F-score in the English datasets.

To evaluate ULIED on the Persian language, we use Babelfy\footnote{ http://babelfy.org } as the Baseline, which works based on BabelNet 3.0\footnote{ https://babelnet.org }. In the first step, We run Babelfy on our dataset by public APIs of Babelfy. Babelfy returns all of the BabelNet synsets for each token in the text. Each synset is linked to some sources, including Wikipedia articles. FarsBase knowledge base is constructed from Persian Wikipedia and uses Wikipedia articles to construct its entities. Therefore, although we only get Persian Wikipedia sources for each BabelNet synset and convert it to its corresponding entity in the FarsBase.
Babelfy (despite its multilingual nature) is not expected to perform comparably with ULIED, in Persian, because (in English) Babelfy utilizes additional lexical data sources such as WordNet whereas, in Persian, neither of Babelfy and ULIED have access to additional lexical resources or knowledge resources to Wikipedia. Indeed, it is the reason for the high difference of ULIED F-score compared to Babelfy, in the Persian language. Babelfy API does not return entity candidates in the results; Thus, comparing the reported recall rate with the ParsEL is not rational, and predictably the recall of our baseline method is lower than ULIED.
It should be noted that it was not possible to compare ULIED with DBpedia-SpotLight. DBpedia-SpotLight covers specific languages for entity disambiguation, and Persian is not one of those languages.

\section{Conclusion and Future Trends}
In this paper, we presented an approach for UnSupervised, Language-Independent Entity Disambiguation, dubbed as ULIED. ULIED utilizes only the link-graph of Wikipedia pages (for each language) and the raw text of their corresponding articles.

ULIED is compared to Babelfy and DBpedia-Spotlight as the state-of-the-art of unsupervised and language-independent entity disambiguation methods. For the English datasets, ULIED F-score is almost similar to (and in some cases better than) Babelfy and almost always better than Spotlight. However, in the Persian dataset\footnote{http://farsbase.net/ParsEL.html} (proposed in the conference version of this paper \cite{fakhrian2019parsel}), it meaningfully outperforms the Babelfy F-score. This expected superiority is because (in English) Babelfy utilizes additional data resources (e.g., WordNet), and its performance comes down, for the cases in which such additional lexical and knowledge resources are absent or are not accessible.

ULIED is suggested for the applications in which training data is not available or costly to produce as an unsupervised method. However, ULIED is not suggested to the problems for which large enough annotated entity linking corpora are available, for which the supervised methods are the best solution.

As the future work of this study, we plan to improve the proposed method to be an end-to-end approach, by focusing on the mention detection sub-task of entity linking. 

Additionally, using phrase embedding can also improve the proposed method, especially for the context similarity detection phase. As another future work, the strategy of weight aggregation is expected to be upgraded. Moreover, the proposed method is expected to be evaluated in other languages, especially in the multilingual entity linking datasets.

\section{Acknowledgments}The authors certify that they have NO affiliations with or involvement in any organization or entity with any financial interest, or non-financial interest in the subject matter or materials discussed in this paper.

\bibliographystyle{unsrt}
\bibliography{ULIED}

\typeout{get arXiv to do 4 passes: Label(s) may have changed. Rerun}
\end{document}